%% file: main.tex
\documentclass[runningheads]{llncs}
\usepackage[T1]{fontenc}
\usepackage[numbers]{natbib}

\usepackage{graphicx}
\usepackage{multirow}
\usepackage{xcolor}
\usepackage{makecell}
\usepackage{rotating}
\usepackage{amssymb}

\usepackage{algorithm}
\usepackage{algorithmic}
\usepackage{amsmath}
\begin{document}
\title{Prompt2DeModel: Declarative Neuro-Symbolic Modeling with Natural Language}
\titlerunning{Prompt2DeModel}
%
\author{
    \textbf{Hossein Rajaby Faghihi}\textsuperscript{\rm 1}, \textbf{Aliakbar Nafar}\textsuperscript{\rm 1},
    \textbf{Andrzej Uszok}\textsuperscript{\rm 2},
    \textbf{Hamid Karimian}\textsuperscript{\rm 1}, and 
    \textbf{Parisa Kordjamshidi}\textsuperscript{\rm 1}
}
\authorrunning{Hossein Rajaby Faghihi et al.}
%
%
\institute{Michigan State University \and
Florida Institute for Human and Machine Cognition\\
    \texttt{\{rajabyfa, nafarali, karimian, kordjams\}@msu.edu, auszok@ihmc.org}
}
\maketitle              
\begin{abstract}
This paper presents a conversational pipeline for crafting domain knowledge for complex neuro-symbolic models through natural language prompts. It leverages large language models to generate declarative programs in the DomiKnowS framework. The programs in this framework express concepts and their relationships as a graph in addition to logical constraints between them. The graph, later, can be connected to trainable neural models according to those specifications. Our proposed pipeline utilizes techniques like dynamic in-context demonstration retrieval, model refinement based on feedback from a symbolic parser, visualization, and user interaction to generate the tasks' structure and formal knowledge representation. This approach empowers domain experts, even those not well-versed in ML/AI, to formally declare their knowledge to be incorporated in customized neural models in the DomiKnowS framework.

\keywords{Neuro-Symbolic AI  \and Code Generation \and Constraint Integeration \and Declarative Programming \and Language to Code \and Generative AI \and Large Language Models.}
\end{abstract}
\section{Introduction}
With the rise of large language models~(LLM), the community is getting closer to the dream of natural language programming interfaces~\cite{nijkamp2022codegen,roziere2023code}. Nonetheless, developing systems capable of seamlessly and accurately interpreting and executing complex programming tasks using natural language remains a challenging goal.
Conversely, declarative programming has endeavored to express programs at the high-level problem specifications~\cite{abe0560f66034ae3893128ebc0e8197e}. Within this programming paradigm, emphasis on the logic of what needs to be accomplished, as opposed to the intricacies of implementation, reduces the gap between natural and declarative languages. Inspired by this idea, our research envisions a natural language interface for a \textit{declarative learning-based programming} framework~\cite{Roth99c,Kordjamshidi2018SystemsAA,10.5555/2832415.2832505}. Within the interface, developers articulate their learning tasks using natural language, where the underlying system translates them into code in the declarative language of DomiKnowS~\cite{faghihi2021domiknows}. DomiKnowS further establishes connections between symbolic knowledge and neural learning components (currently designed in PyTorch), facilitating the integration of domain knowledge into deep learning models.

Given the rise of LLMs and their ability to be universal computational models,  the need for customized models can be debatable. However, we follow two main arguments that drive the need for developing customized learning-based models: 1) Despite their impressive performance across various tasks~\cite{zhao2023survey}, Large Language Models~(LLMS) trail behind smaller, domain-specific models~\cite{wu2023exploring, viswanathan2023prompt2model, kocon2023chatgpt,mirzaee2023disentangling}, and 2) incorporating domain knowledge into models, apart from the performance gains,  improves the interpretability and reliability of the models by adhering to domain constraints~\cite{pan2023logic, RajabyFaghihi_Gluecons, yang2023coupling}. Imposing domain knowledge has the potential to leverage explicit alignment of the models with safety, fairness, and ethical constraints~\cite{hoernle2022multiplexnet}.



In this work, we leverage the progress on specialized frameworks for knowledge integration~\cite{faghihi2021domiknows,ahmed2022pylon,manhaeve2018deepproblog,huang2021scallop}, particularly those with declarative interfaces~\cite{kordjamshidi2022declarative} and the advancements in tailoring neural model architectures using LLMs~\cite{viswanathan2023prompt2model}.

Prior research on tailoring deep learning models with natural language prompts has either focused on fine-tuning simple text-to-text models~\cite{viswanathan2023prompt2model} or composing complex architectures during inference~\cite{shen2023hugginggpt}. However, our proposal facilitates the development of complex deep-learning models and enables further fine-tuning them on the task-specific data within one framework.

Adapting LLMs to generate DomiKnowS declarative programs presents several challenges. These include limited resources for fine-tuning~(There are few training samples available in the declarative language of DomiKnowS, unlike popular programming languages such as Python or frameworks such as PyTorch.), the requirement to adjust LLMs to a new coding style specific to DomiKnowS, the complexity of translating domain knowledge into First-Order Logic~(FOL) statements~\cite{han2022folio}, let alone the specific DomiKnowS' Python-based FOL language, and the potential lack of user familiarity with the DomiKnowS language, hindering meaningful feedback in a human-in-the-loop process.

To tackle these issues, we propose an interactive pipeline where we utilize an underlying LLM with techniques such as prompt templates for user interactions, dynamic few-shot in-context learning, intermediary mapping of natural language~(NL) to FOL statements, and iterative refinement based on feedback from symbolic semantic/syntactic verification functions to generate the final declarative modeling code.
The pipeline's interface \footnote{To access the demo, use \url{https://hlr-demo.egr.msu.edu/} (Please email authors for login access.). For demonstration, see \url{https://youtu.be/9q9PvH3dJKE}} can be accessed online.



\section{Related Research}

We aim to facilitate the development of neuro-symbolic models leveraging background knowledge expressed via logical constraints. Using logical knowledge, explored both during inference~\cite{freitag2017beam,scholak2021picard,guo2021inference,rajaby-faghihi-kordjamshidi-2024-consistent} and training~\cite{hu2016harnessing,nandwani2019primal,xu2018semantic} of neural networks, has been encapsulated in other libraries like DeepProbLog~\cite{manhaeve2018deepproblog}, PyLon~\cite{ahmed2022pylon}, and Scallop~\cite{huang2021scallop}. DomiKnowS~\cite{faghihi2021domiknows} provides a declarative interface for defining knowledge and computational units, allowing seamless integration of knowledge using various underlying techniques. While DomiKnowS represents a significant advancement, our proposal seeks to enhance the declarative interface by enabling the direct use of task descriptions in natural language form.

The existing endeavors in code generation~\cite{chowdhery2022palm,touvron2023llama} deploy large models in few-shot or zero-shot settings~\cite{zhang2023coder,chen2022codet} or fine-tune them for specific tasks~\cite{nijkamp2022codegen,roziere2023code}. These approaches demand data resources similar to the target task in pre-training/fine-tuning, which are inaccessible to us. Our work emphasizes using natural language interfaces to generate code that supports the development of neuro-symbolic models instead of conventional code in popular programming languages.

Recent research on generating neural architecture from natural language prompts has focused on either composing models during inference~\cite{shen2023hugginggpt}, generating training data~\cite{yu2023alfred}, or training straightforward LLM architectures without extensive modeling efforts~\cite{viswanathan2023prompt2model}. In contrast, we map natural language prompts to formal representations to create complex declarative neuro-symbolic models.


Our proposal aligns with recent research on the trajectory of deriving formal representations from language models, where 
recent research has delved into such translation to formalism suitable for underlying engines for logical inference and constraint optimization domains~\cite{pan2023logic}, as well as guided generation for consistent reasoning~\cite{poesia2023certified} or causal~\cite{jin2023cladder} and probabilistic reasoning~\cite{nafar2024probabilistic}. However, we propose an interactive pipeline for extracting representations, facilitating the development of neural architectures intricately connected to symbolic concepts. Our task includes an additional complexity, as the target formalism is specific to DomiKnows and not seen during LLMs' pretraining.

Lastly, our work employs techniques for sampling LLMs' output, emphasizing that correct response can be obtained with a sufficient number of samples~\cite{ni2023lever}. Techniques like majority voting, post-pruning, and filtering by test cases have enhanced accuracy in large-scale sampling~\cite{chen2021evaluating,austin2021program,cobbe2021training}. Due to the dynamic nature of tasks and neural models, defining test cases is impossible for our use case. Instead, we employ an iterative feedback system that provides execution errors to the language model to reflect on. In contrast to self-refinement research~\cite{huang2022large}, our work introduces an external symbolic parser for the semantic evaluation of expected output structures, offering insights for iterative error correction.
\begin{figure}[t]
    \centering
    \includegraphics[width=0.5\linewidth]{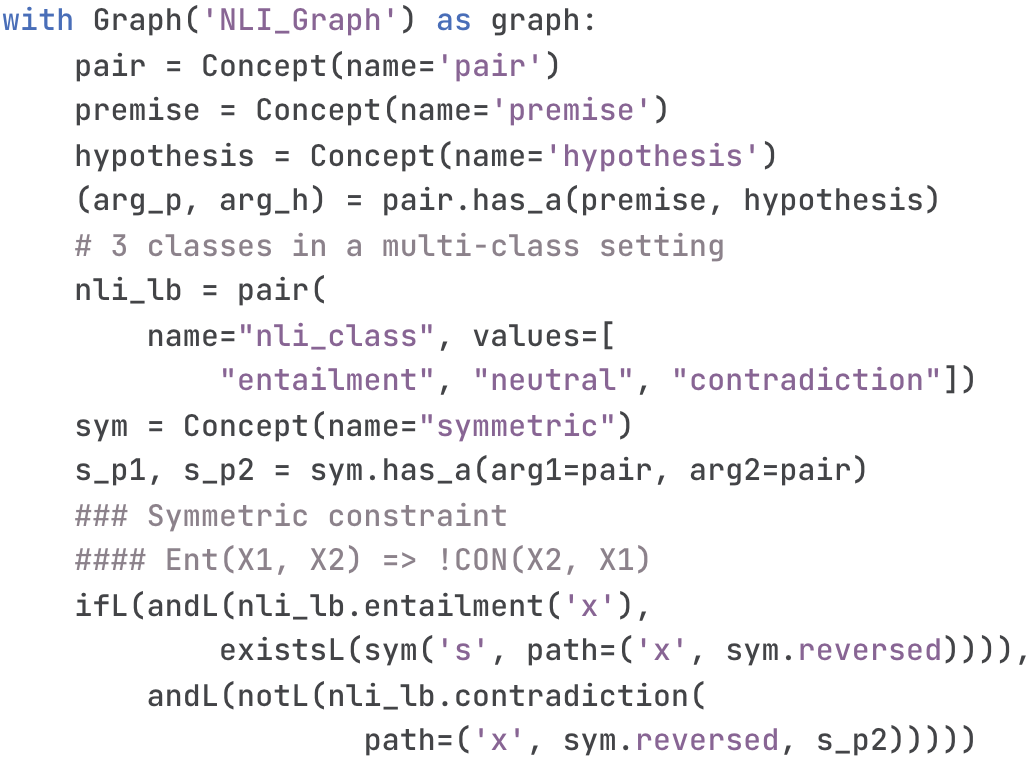}
    \caption{Parts of the concept graph generated for the Natural Language Inference task.}
    \label{fig:sample}
    \vspace{-3mm}
\end{figure}
\begin{figure}[t]
    \centering
    \includegraphics[width=\linewidth]{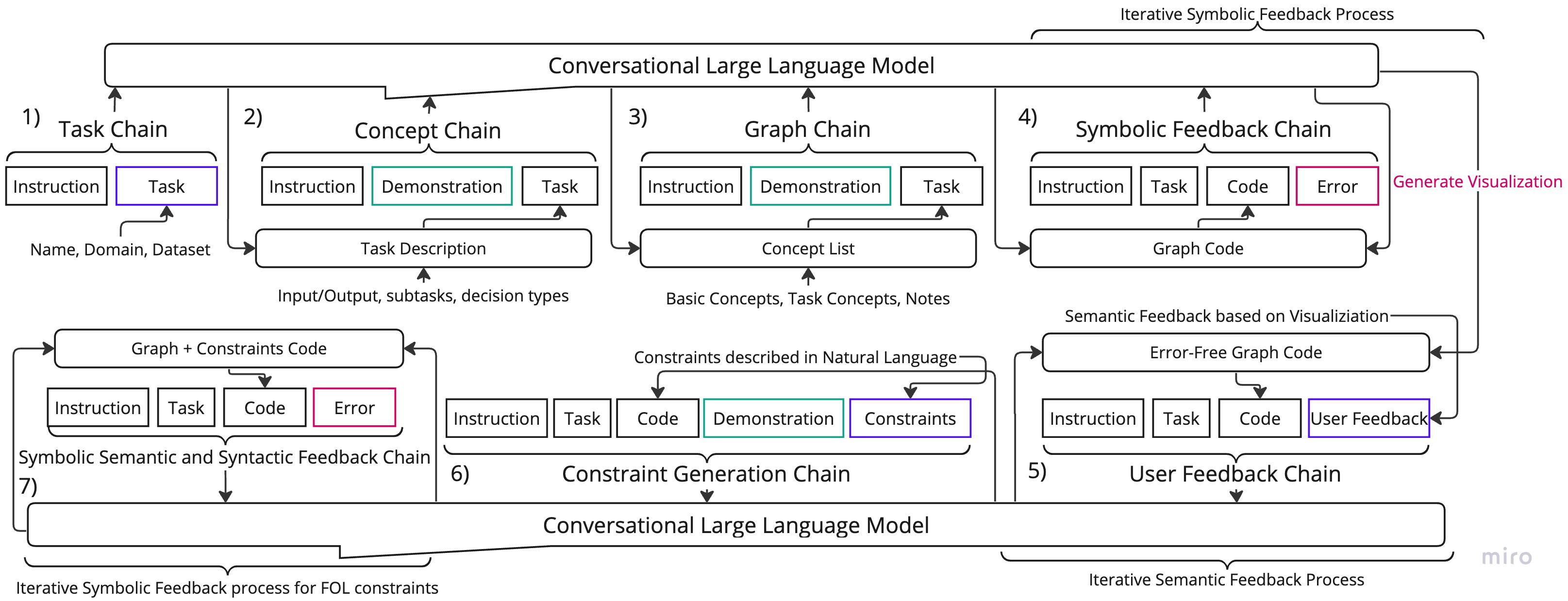}
    \caption{Overview of the pipeline: The purple, green, and red parts represent direct user inputs, dynamic in-context examples, and execution/parser feedback, respectively.}
    \label{fig:pipeline}
    \vspace{-5mm}
\end{figure}
\section{DomiKnowS Interface}
Our tool aims to translate natural language descriptions of structured prediction tasks into corresponding representations within the DomiKnowS framework. DomiKnowS programs include three components: knowledge declaration~(domain/ task structure and constraints), model declaration~(computational units), and program execution~(knowledge integration, learning, and inference). This paper focuses on the knowledge declaration, as it often requires input from domain experts who may lack familiarity with neural architectures or programming languages and benefit significantly from a natural language interface.
The knowledge declaration stage results in a concept graph that represents the structure of tasks and incorporates domain-specific logical constraints. A snippet of code in DomiKnowS, illustrating the representation of the Natural Language Inference (NLI) task graph structure and logical constraints, is provided in Figure \ref{fig:sample}.

\subsection{Graph Structure}

The graph illustrates the input-output structure and dependencies of the task. It includes nodes representing different concepts and edges denoting their relationships. In NLP tasks, input concepts consist of sentences, paragraphs, phrases, words, and tokens. Vision tasks involve concepts related to images and bounding boxes. In structured prediction, the output concepts are task labels. Each decision concept needs to be anchored to an input concept. For example, the concept of \textit{sentence\_class} is connected to the \textit{sentence} concept, and \textit{named entity} tags are associated with the \textit{phrase} concept. DomiKnowS uses ``is\_a'' to represent connections between a concept and its parent, ``contains'' for relations between a concept and its same-type children, and ``has\_a'' for many-to-many relations.

\subsection{Logical Constraints}

DomiKnowS introduces a domain-specific Python-based language that employs concepts, edges, variables, and logical operators to declare dependencies in first-order logic between concepts, which defines the tasks' domain knowledge in the form of constraints~(See Fig \ref{fig:sample} for a sample code.). Significantly, the syntax for specifying first-order logic constraints in DomiKnowS is not observed during the pre-training of large language models, presenting a difficulty in generating outputs that adhere to this format in a zero or few-shot setting.

\section{Natural Language to Knowledge Declaration} 
Our interactive interface employs prompt templates from the LangChain framework\footnote{https://www.langchain.com/} and GPT-3.5-turbo. At each stage, user input fills in a predefined template that the model consumes to generate the information needed for subsequent steps. We adjust the conversation history to prevent excessive context accumulation, especially where we have an iterative process.

Figure~\ref{fig:pipeline} provides an overview of the pipeline. The model can consume a series of underlying natural language instructions, in-context demonstrations, and execution feedback, in addition to the user input and information from prior interactions at each step. 
The process begins by gathering basic information, such as the task's name, domain, and dataset name~(see step 1 in the figure). Subsequently, the model formulates a clear task description, including the task's input/output and decision types. Users have the flexibility to modify this description. Next, the model compiles a list of concepts~(nodes in the graph) representing the task's structure~(the input/output concepts, decision types, and more; See step 2). Users can modify items on this list, which is the foundation for generating an initial concept graph~(in Python; See step 3).

\begin{figure}[h]
    \centering
    \includegraphics[width=0.5\linewidth]{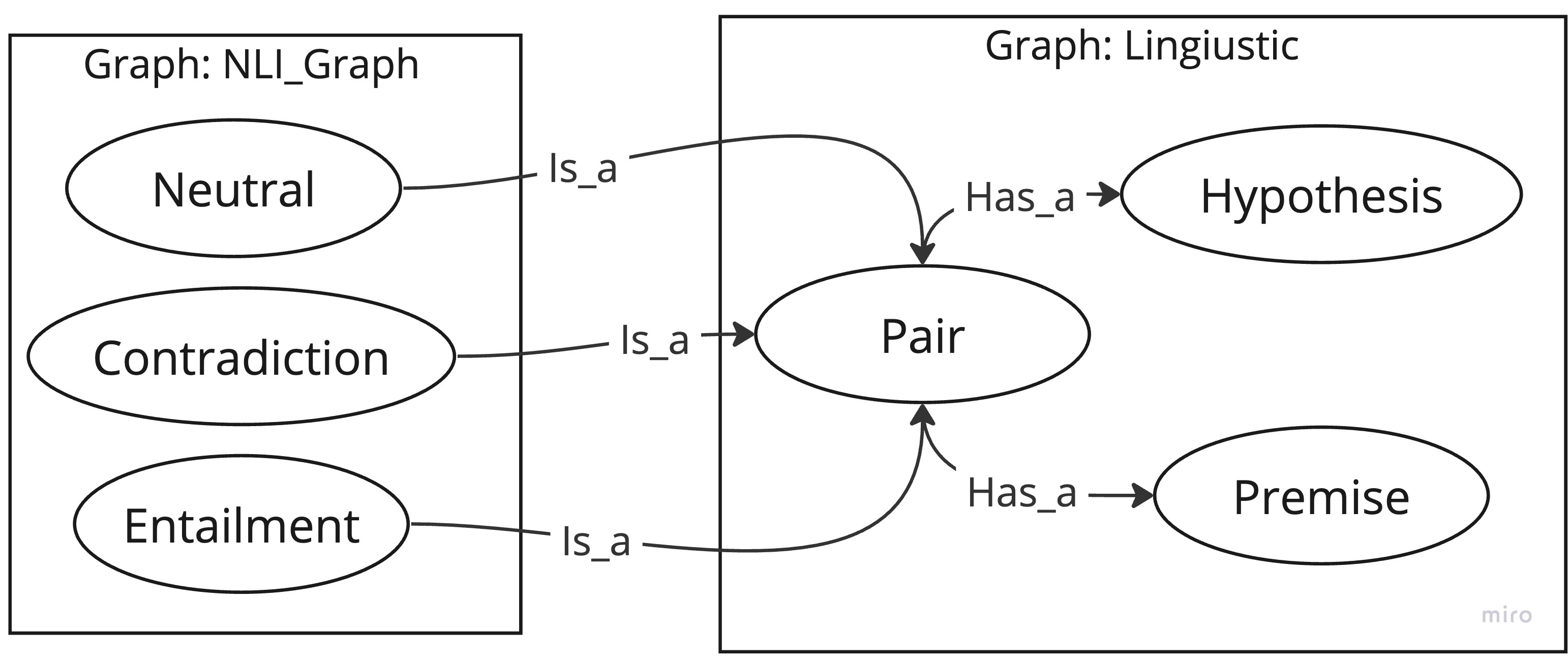}
    \caption{The visualized graph for the NLI task.}
    \label{fig:graph}
    \vspace{-4mm}
\end{figure}
The generated code might not be accurate/executable due to the model's lack of familiarity with DomiKnowS' language syntax. Therefore, we implement a loop for parsing the code and detecting the errors while providing the LLM with feedback for refinement. At each step, the model receives errors from the code's parse and execution, which evaluates the graph structure and uses this feedback to refine the code~(Step 4). We also seek user feedback once confident in the code's quality. As users need not be familiar with DomiKnowS' coding structure, they provide feedback based on the visualization of the code, representing the nodes and edges in the graph~(Step 5). Figure \ref{fig:graph} visualizes the graph for the NLI task.
Subsequently, users provide the task constraints in natural language. The model aims to translate these expressions into DomiKnowS' Python-based FOL language, which differs somewhat from standard FOL syntax~(see step 6). We employ techniques such as dynamic in-context learning, intermediary FOL mapping, and iterative refinement to assist the model in this translation~(step 7).
Here, we briefly discuss some of the important techniques used in the pipeline.

\noindent\textbf{Sampling Strategy} We draw multiple samples from the underlying LLM at each step and select the most accurate response through user feedback or automated metrics, such as error counts from the execution. Additionally, if some samples do not have any errors generated by the symbolic processing engine, we prune out the remaining erroneous samples after a certain amount of iterations.

\noindent\textbf{Dynamic In-Context Retrieval} 
Our pipeline involves in-context demonstrations for various tasks in the NLP and vision domains at each step. To enhance model execution speed and minimize noise from irrelevant examples, we selectively include a subset of in-context demonstrations relevant to the target task. We achieve this by creating a vector database of in-context demonstration representations using the OpenAI Embedding service and identifying the most relevant ones for inference based on cosine similarity. This approach reduces computational costs and noise in LLM's generation.

\noindent\textbf{Iterative Symbolic Refinement} We propose an iterative refinement process in response to LLMs' inability to execute code directly to ensure its sanity. Leveraging the representations of structure and constraints in DomiKnowS, we develop syntactic and semantic evaluation parsers to provide feedback. This symbolic parser generates graph structure and constraints, pinpointing semantic and syntactic errors. The parser is a custom-designed algorithm that processes the written code in DomiKnowS and finds errors and inconsistencies in its logic and syntax. The parser's feedback serves as instructions for the LLM to refine its initial answer, addressing the challenge of limited prior exposure to DomiKnowS syntax and ensuring accurate responses.

\noindent\textbf{FOL Mapping} 
To enhance constraint generation precision, we utilize an intermediary FOL mapping process. The model translates natural language descriptions into FOL statements and then converts them into DomiKnowS FOL syntax, leveraging the model's exposure to FOL statements during pre-training.  Logical predicates/arguments are symbolically extracted and included in the input to help in the prediction. FOL mapping assists LLMs in capturing semantics through FOL representations, enabling users to validate constraints based on these statements, assuming perfect alignment between the two representations.

\section{Evaluation and Analysis}
\label{sec:evaluation}
We assessed our method for various tasks in NLP, Vision, and Constraint Satisfaction Problems~(CSP) using both automatic~(similarity-based) and human judgments.
\subsection{Automatic Evaluation}
We devise a set of metrics customized for each step of the process to evaluate the expected output at each stage. This evaluation covers a benchmark of 14 tasks, spanning Entity/Relation Extraction~\cite{tjong-kim-sang-de-meulder-2003-introduction}, Hierarchical Classification~\cite{cifar100}, Sudoku, Procedural Reasoning~\cite{faghihi2023role,faghihi2021time}, Causal Reasoning~\cite{tandon-etal-2019-wiqa}, Natural Language Inference~\cite{bowman2015large}, and more. At each stage, we compare with the ground truth input from the dataset for evaluation. A distinct evaluation approach for end-to-end mode, incorporating human intervention, will be discussed in the next section.

\input{tables/before_constraints}

\noindent\textbf{Task, Concept, and Graph Generation}
Table \ref{tab:before_constraints} reports the results using automatic metrics that evaluate task description, concepts, and initially generated and corrected graphs. The outcomes demonstrate the success of our proposed tool in generating the information at each step. The large number of error-free tasks after refinement and the higher performance of the model using sampling indicate the effectiveness of both of these techniques in enhancing the model's performance. 
Notably, generating the concept list shows a higher error rate, especially in tasks like image classification in CIFAR-100~\cite{cifar100}, where decision space is large, and the model often falls short by adding `etc.' to the label set instead of listing all classes. The limited differences in predicted nodes and edges compared to the ground truth suggest that human intervention can effectively address graph errors by instructing the model to add or remove small components. Although dynamic retrieval can help reduce the cost and increase the speed of the process, it slightly hurts the model's performance at these phases. 

\input{tables/constraints}

\noindent\textbf{Constraints Generation}
Table \ref{tab:constraints} displays experimental results for evaluating the model's ability to generate semantically/syntactically correct logical constraints within the graph structure and Python syntax of DomiKnowS.
As indicated by the high percentage of resolved errors~(\textit{\%S} in \textit{\%E in Resolved E}) detected by our custom-designed parser, it plays a significant role in improving program accuracy. This is accomplished through the feedback loop, by providing clear descriptions of both syntactic and semantic errors.
It is notable that a small sampling factor~(3 samples) enhances model accuracy, and FOL mapping serves as a valuable intermediary layer, reducing the need for iterative constraint adjustments. Manual evaluation reveals that while most generated constraints~(even in erroneous tasks) are semantically accurate, issues arise in adapting to the DomiKnowS language due to limited prior exposure of the LLM to its specific syntax. While dynamic retrieval aids in direct constraint generation, it hinders performance using FOL due to the lack of exposure to FOL-to-DomiKnowS mappings, which encourages the use of unsupported FOL operations in DomiKnowS.


\subsection{Human Evaluation}
\label{sec:human}
The human evaluation assesses two key aspects of our demo. Firstly, it measures the comparative ease and utility of our proposed demo in contrast to the conventional method of navigating the documentation and manually crafting programs within the Python package of the DomiKnowS library. Secondly, the evaluation provides insights into the accuracy of each step in the interactive process, detailing the quantity and nature of user interventions required for satisfactory performance across diverse tasks. Additionally, we aim to assess whether the proposed tool facilitates users unfamiliar with the DomiKnowS programming language in validating the model's responses.

\subsubsection{Interface V.S. Coding}

We assess user experiences in crafting structured prediction tasks in DomiKnowS with two volunteer groups~(5 people per group). Each group received a 20-minute introduction to structured prediction and a high-level overview of DomiKnowS. The first group used the demo interface, while the second group, with access to DomiKnowS documents and examples, manually composed concept graphs and logical constraints. Tasks included hierarchical image classification, sentiment analysis, and entity mention and relation extraction. 

\noindent\textbf{Groups' Sub-tasks:} Group 1: Describe the task and interact with the interface. Group 2: Read the documentation and examples; Write the code.

\noindent\textbf{Average Time of Implementation: } Group 1: 20 minutes, Group 2: 1 hour

\noindent\textbf{Time Splits:} Group 1: 60\% user input - 40\% model output.
Group 2: 35\% documents - 65\% coding

\noindent\textbf{Findings:} Human effort using the interface is reduced 5 times. Opting for the local LLM version instead of the API can further accelerate final code generation. Notably, the second group often sought additional guidance on task modeling, while the tools integrated into the demo interface effectively guided the first group.
Additionally, we instructed the second group to revisit the task using the demo interface, allocating them a 10-15 minute timeframe. The consensus within this group was that the demo interface provided a significantly superior and more user-friendly experience.

\subsubsection{Interactive Setting}
To evaluate the demo in an end-to-end interactive setting, we asked two volunteers to implement a total of \textbf{seven} tasks while recording their every interaction with the interface. 

\noindent\textbf{Duration: }The average time for the process was 17 minutes. The most time-consuming part was the constraint generation, taking more time for both the user to write constraints and the model to generate them~(averaging 150 seconds for model responses).

\noindent\textbf{Task Descriptions:} 
In 6 out of 7 tasks, the user removed only additional and unnecessary information. In one task, the user extended the decision set by replacing the word `etc.' with actual labels.

\noindent\textbf{Concept List:} In 4 out of 7 tasks, one sample was correct, while in the remaining tasks, the user had to remove or add less than two concepts to the list.

\noindent\textbf{Concept Graph:} 
In 4 out of 7 tasks, the correct graph was generated without user intervention. In 2 out of 7 tasks, one interaction was needed. In 1 out of 7 tasks, the user interacted five times to remove a wrongly included relationship between concepts where multiple relationships existed. The visualization tool notably provided users with all the necessary information to evaluate the graph.

\noindent\textbf{Constraints:}
Most of the constraints were both semantically and syntactically correct. In cases with erroneous results, the verification process could capture the error but not resolve it correctly. Errors were mainly due to using syntax close to FOL for operations not directly supported in DomiKnowS, like equality between multiple variables. Remarkably, in cases such as constraints in a sudoku table, the model detected similar patterns and used for-loops to implement multiple constraints with similar logical structures but variant logical predicates.

\section{Conclusion}
We proposed a natural language interface for declarative programming to facilitate the design of neuro-symbolic models by domain experts.
We exploit LLMs and prompt engineering techniques to convert natural language into declarative code, written in a specific syntax for the \textit{DomiKnowS} neuro-symbolic modeling framework. It includes the structure of concepts and relationships related to the problem, as well as formal logical constraints expressing domain knowledge.
Building on top of these declarations, \textit{DomiKnowS} integrates domain knowledge with deep models of any kind designed in PyTorch. This framework is a step forward in making the design of deep models accessible to domain experts, allowing them to inject their domain knowledge into the model using natural language.

\section*{Acknowledgments}
This project is supported by the National Science Foundation (NSF) CAREER award 2028626 and partially supported by the Office of Naval Research
(ONR) grant N00014-20-1-2005 and  grant N00014-23-1-2417. Any opinions,
findings, and conclusions or recommendations expressed in this material are those of the authors and do not necessarily reflect the views of the National
Science Foundation nor the Office of Naval Research.
Thanks to our volunteers, Tanawan Premsri, Juan Castro-Garcia, Danial Kamali, and Max Reuter, who helped us in the pipeline evaluation process.

\bibliography{anthology,custom,previous}
\end{document}

%% file: tables/before_constraints.tex
\begin{table*}[]
\vspace{-4mm}
\centering
\footnotesize
\caption{Summary of automatic evaluation results for generating task components (description, concept list, initial graph, and post-fix graph). In full in-context, 4 demonstrations are used, while Dynamic has 1. $E$ stands for Error. $n/N$ indicates successful graph parsing of $n$ out of $N$ tasks without structural issues. $x$ N, $y$ E represents average $x$ differences in nodes and $y$ differences in edges compared to the ground truth. When using more than 1 sample, the result of the best sample is reported, mimicking human intervention in selecting the optimal answer. $^*$ the values are calculated by BertScore.}
\begin{tabular}{|c|c|c|c|cc|c|}
\hline
\multirow{2}{*}{In-Context} & \multirow{2}{*}{\# Samples} & \multirow{2}{*}{Task Desc$^{*}$} & \multirow{2}{*}{Concept Diff} & \multicolumn{2}{c|}{Initial Graph}             & Refined Graph \\ \cline{5-7} 
                            &                            &                                               &                                    & \multicolumn{1}{c|}{E-Free} & Diff         & E-Free    \\ \hline
Full                        & 1                          & 73.9\%                                        & 20.8\%                             & \multicolumn{1}{c|}{12/14}      & 0.8 N, 3.3 E & 14/14         \\ \hline
Full                        & 3                          & \textbf{79.3}\%                                        & \textbf{20.6}\%                             & \multicolumn{1}{c|}{\textbf{13/14}}      & \textbf{0.76 N, 2 E}  & 14/14         \\ \hline
Dynamic                     & 1                          & 73.9\%                                        & 26.8\%                             & \multicolumn{1}{c|}{10/14}      & 2.1N, 2.6 E  & 14/14         \\ \hline
Dynamic                     & 3                          & \textbf{79.3}\%                                        & 26.8\%                             & \multicolumn{1}{c|}{\textbf{13/14}}      & 1.5 N, 2.1 E & 14/14         \\ \hline
\end{tabular}
\label{tab:before_constraints}
\end{table*}

%% file: tables/constraints.tex
\begin{table*}[]
\vspace{-2mm}
\small
\centering
\caption{Results for constraint generation and the ratio of error types in resolved/unresolved errors during iterative feedback. Four in-context demonstrations are available in Full and one in Dynamic. $E$ stands for Error. `$i$ D, $j$ I' denotes $i$ tasks are correct without the need for feedback and $j$ are fixed after the iterative feedback. $i\%$ P, $j\%$ S indicates that $i\%$ of resolved/unresolved errors are Python errors, while $j\%$ are errors caught and reported in a custom prompt with our symbolic parser. In multi-sample settings, an error-free task has at least one sample with no errors.}
\begin{tabular}{|c|c|c|c|c|c|}
\hline
Setting                 & In-Context               & \# Samples & E-Free Tasks & \% E in Resolved E & \% E in Unresolved E \\ \hline
\multirow{4}{*}{Direct} & \multirow{2}{*}{Full}    & 1        & 2 D, 6 I         & 53\% P, 47\% S  & 79\% P, 21\% S \\
                        &                          & 3        & 4 D, 9 I         & 48\%P, 52\% S   & 66\% P, 34\% S \\ \cline{2-6} 
                        & \multirow{2}{*}{Dynamic} & 1        & 5 D, 5 I         & 70\% P, 30\% S  & 70\% P, 30\% S \\
                        &                          & 3        & 6 D, 7 I         & 41\% P, 59\% S  & 64\%P, 36\%S   \\ \hline
\multirow{4}{*}{FOL}    & \multirow{2}{*}{Full}    & 1        & 8 D, 0 I         & 90\% P*, 10\% S & 80\% P, 20\%S  \\
                        &                          & 3        & 13 D, 0 I        & 77\%P, 23\% S   & 82\%P, 18\%S   \\ \cline{2-6} 
                        & \multirow{2}{*}{Dynamic} & 1        & 5 D, 2 I         & 65\%P, 35\% S   & 80\% P, 20\% S \\
                        &                          & 3        & 10 D, 1 I        & 65\%P, 35\%S    & 75\%P, 25\% S  \\ \hline
\end{tabular}
\label{tab:constraints}
\end{table*}